\documentclass[11pt,a4paper]{article}
\usepackage{coling2016}
\usepackage{times}
\usepackage[english]{babel}
\usepackage[utf8x]{inputenc}
\usepackage[T1]{fontenc}
\usepackage{graphicx}
\usepackage{morefloats}
\usepackage{arydshln}
\usepackage{multirow}
\usepackage{dashbox}
\usepackage{roundbox}
\usepackage{listings}
\usepackage{url}

\title{Word2Vec vs DBnary: Augmenting METEOR using Vector Representations or Lexical Resources?}

\author{\\
  \\
  \\
  $^{1}$LIG -- Univ. Grenoble Alpes  \\ 
  Domaine Universitaire \\
  38401 St Martin d'H\`eres, France \\
  {\tt firstname.lastname}\\{\tt @imag.fr} \\
  \And 
  Christophe Servan$^{1,2}$, Alexandre Bérard$^{1}$, Zied Elloumi$^{1,3}$,\\ \textbf{Hervé Blanchon$^{1}$ \& Laurent Besacier$^{1}$} \\
  \\
  $^{2}$SYSTRAN \\
  5 Rue Feydeau \\
  75002 Paris, France \\
  {\tt firstname.lastname}\\{\tt @systran.fr} \\
  \And
  \\
  \\
  \\
  $^{3}$LNE  \\
29 Avenue Roger Hennequin \\
78190 Trappes, France 
\\
  {\tt firstname.lastname}\\{\tt @lne.fr} 
}

\date{}

\begin{document}

\maketitle

~\\

\begin{abstract}
This paper presents an approach combining lexico-semantic resources and
distributed representations of words applied to the evaluation in machine translation (MT).
This study is made through the enrichment of a well-known MT evaluation metric: METEOR.
This metric enables an approximate match (synonymy or morphological similarity) between an automatic and a reference translation.
Our experiments are made in the framework of the \textit{Metrics} task of WMT 2014. 
We show that distributed representations are a good alternative to lexico-semantic resources for MT evaluation and they can even bring interesting additional information. 
The augmented versions of METEOR, using vector representations, are made available on our \textit{Github} page.\newline \newline
\end{abstract}


\section{Introduction}
\blfootnote{This work is licensed under a Creative Commons Attribution 4.0 International Licence. Licence details:\\\url{http://creativecommons.org/licenses/by/4.0/}}
Learning vector representations of words using neural networks has generated a strong enthusiasm in the NLP research community. 
In particular, many contributions were proposed after the work of \cite{MikolovICLR2013,MikolovNIPS2013,Mikolov2013} on training word embeddings.
The main reasons for this strong interest are: the proposal of a simple and efficient neural architecture to learn word vector representations, the availability of an open source tool \textit{Word2Vec}\footnote{http://word2vec.googlecode.com/svn/trunk/} and the rapid structuring of a user community\footnote{https://groups.google.com/d/forum/word2vec-toolkit}. 
Later on, several contributions have extended the work of Mikolov on word vectors to  phrases (sequences of words) \cite{MikolovNIPS2013,le_distributed_2014}   and to bilingual representations  \cite{luong_bilingual_2015}. 
All these vector representations capture similarities between words, phrases or sentences at different levels (morphological, semantic).

However, although these representations can be semantically informative, they do not exactly replace fine-grained information available  in lexical-semantic resources such as \textit{WordNet} \cite{Fellbaum1998},  \textit{BabelNet} \cite{DBLP:conf/acl/NavigliP10}, or \textit{DBnary} \cite{Serasset2012}. Such lexical resources are also more easily interpretable by humans as shown in \cite{Panchenko2016}, but their construction is costly while word embeddings can be trained  \textit{ad infinitum} on any monolingual or bilingual corpora. 

In short, both approaches (lexical resources and word embeddings) have their pros and cons. However, few
studies have attempted to compare and combine them.  Pioneering work of \newcite{FaruquiDJDHS14}  proposed to refine representations learning using
lexical resources. The idea is to force  words connected in the lexical network, to have a close representation (for example through a synonymy link). The technique
proposed is evaluated on several benchmarks (word similarity, sentiment analysis, finding of synonyms). More recently, \newcite{Panchenko2016} and \newcite{rothe-schutze2015ACL-IJCNLP} extended word embeddings to sense embeddings and tried to compare them to lexical synsets.

 \textbf{Contributions:} this article attempts to  review the contribution of vector representations to
measure sentence similarity. We compare them with similarity measures based on lexical resources such as \textit{WordNet} or \textit{DBnary}.
Machine Translation (MT) evaluation was identified as a particularly interesting application to investigate, since MT evaluation is still an open problem nowadays. More precisely, we propose to augment a well known MT evaluation metric (METEOR \cite{Lavie2005}) which allows an approximate matching
(through synonymy or morphological similarity) between MT hypothesis and reference. The augmented versions of METEOR proposed (using word embeddings, lexical resources or both) allow us to objectively compare the contribution of each approach to measure sentence similarity. For this,  correlations between METEOR and human judgements (of MT outputs) are measured within the framework of WMT 2014 \textit{Metrics} task. The code of the augmented versions of METEOR is also provided on our \textit{Github} page\footnote{\url{https://github.com/cservan/METEOR-E}}.

\textbf{Outline:} in section \ref{sec:EA} (Related Work),  we quickly present METEOR, lexical resources and word embeddings. Section \ref{sec:eval} presents our propositions to augment METEOR in order to conduct a fair comparison between lexical resources and vector representations respectively. Section \ref{sec:correlation} presents our experiments made within the framework of WMT 2014, as well as quantitative and qualitative analyses. Finally, section \ref{sec:concl} concludes this work and gives some perpectives.

\section{Related Work}
\label{sec:EA}
\subsection{An automatic metric for MT evaluation: METEOR}
\label{subsec:meteor}
\subsubsection{The origins}

METEOR was proposed to compensate BLEU's and NIST's weaknesses \cite{Papineni2002,Doddington2002}.
In short, METEOR was created to better correlate with human judgements by using more than word-to-word alignments between a hypothesis and some references. 
The alignment is made according to three \textit{modules}: the first stage uses exact match between word surface forms (\textit{Exact} module), the second one compares word stems (\textit{Stems} module) and the third one uses synonyms (\textit{Synonym} module) from a lexical resource such as WordNet (available for English only in METEOR).

One contribution of this paper is to propose an alternative to \textit{Stems} and \textit{Synonym} modules: our proposed add-on will be called \textit{Vectors} module later on.

\subsubsection{Recent extensions of METEOR}

METEOR-NEXT \cite{Denkowski2010a} was proposed to better correlate with HTER (Human-targeted Translation Edit Rate -- \textit{HTER} 
\cite{Snover2006}).
HTER is a semi-automatic post-editing based metric, which measures the edit distance between a hypothesis and a reference.
METEOR-NEXT proposes to go further than just word-to-word alignment by using phrase-to-phrase alignments.
For this, phrase databases were created for 
several languages like English \cite{Snover2009a}, German, French or Czech 
\cite{Denkowski2010}. More recently, another version called \textit{METEOR Universal} used bitexts to extract paraphrases \cite{Denkowski2014}.

METEOR was also extended by using Word Sense Disambiguation (WSD)
techniques \cite{Apidianaki2015}. The authors used \textit{Babelfly} 
\cite{Moro2014} for several langage pairs (translation from French, 
Hindi, German, Czech and Russian to English). A better correlation with human judgement at segment level was observed using WSD in METEOR.

Finally, to extend the use of \textit{Synonym} module to target languages others than English, \newcite{Elloumi2015} proposed to replace WordNet by
DBnary \cite{Serasset2012}. The new target languages equipped with a \textit{Synonym} module were French, German, Spanish, Russian and English.

\subsection{Lexical resources}
\subsubsection{WordNet}

WordNet is a well  known lexical resource for English. Created at the University of Princeton \cite{Fellbaum1998}, it is used in several NLP tasks such as Machine Translation, Word Sense Disambiguation, Cross-lingual Information Retrieval, etc. 
WordNet links nouns, verbs, adjectives and adverbs to a set of synonyms called ``synsets''. Each synset represents a specific concept.

Synsets are linked to each other according to semantic, conceptual and lexical relations.
Words with multiple meanings correspond to multiple synsets and meanings are sorted according to their frequency.
WordNet is available in several languages (Arabic, French, etc.) but these versions are not freely available.
In METEOR, only English WordNet is used to match hypothesis and reference words according to their meanings. It contains more than 117,000 synsets.

To extract lemmatized forms, METEOR uses a function called \textit{Morphy-7WN1} which firstly checks special cases in an exception list and secondly uses rules to lemmatize words according to their syntactic class.

\subsubsection{DBnary}

DBnary is a multilingual lexical resource in RDF format \cite{Klyne2004}. This resource has been collected by \newcite{Serasset2012}.
Lexical data are represented using the LEMON vocabulary \cite{McCrae2011}.
Most Part-of-Speech tags are linked with \textit{Olia standards} or \textit{Lexinfo} vocabularies \cite{Chiarcos2015,Cimiano2011} which makes them reusable in many contexts.

DBnary is downloadable or available online through a SPARQL access point.
Lexical data are automatically extracted from Wiktionary, Wikipedia's dictionary for 21 languages\footnote{Bulgarian, Dutch, English, Finnish, French, German, (Modern) Greek, Indonesian, Italian, Japanese, Latin,
Lithuanian, Malagasy, Norwegian, Polish, Portuguese, Russian, Serbo-Croat, Spanish, Swedish and Turkish}.

\begin{table}[h!]
\centering
\scriptsize
 \begin{tabular}{lrrrr}
 \hline
 & \multicolumn{1}{c}{English} & \multicolumn{1}{c}{French} & \multicolumn{1}{c}{Russian} & \multicolumn{1}{c}{German} \\
 \hline
Number of entries &  620 K & 322 K & 185 K & 104 K \\
Number of meanings  &  498 K & 416 K & 176 K & 116 K \\
Number of synsets  &  35 K &  36 K &  31 K &  33 K \\
 \hline
 \end{tabular}
\caption{\label{tab:DBnary}Detail of the data used from DBnary for the languages targeted in this paper. }
\end{table}

Among available lexical data,
one may find 2.9M lexical entries (with parts-of-speech, canonical form for all of them, along
with pronunciations when available and inflected forms for some languages). Lexical entries
are subdivided into 2.5M lexical senses (with their definitions and some usage example).

DBnary also contains more than 4.6M translations going from the 21 extracted sources
languages to more than 1500 different target languages. Additionally, DBnary contains lexicosemantic relations (syno/anto-nyms, hypo/hypero-nyms, etc.).
Table 1 shows the size of the data for languages involved in the experiments later
reported in this paper. Additional figures are available on the DBnary public web site\footnote{\url{http://kaiko.getalp.org/about-dbnary/}}.

Lemmatized forms for DBnary are based on the \textit{TreeTagger} module \cite{Schmid95}, which enables us to find the corresponding synsets.

\subsection{Monolingual and bilingual embeddings}
\label{sec:WE}
\subsubsection{Overview}

Learning word embeddings is an active research area  \cite{Bengio2003,Turian2010,Collobert2011,Huang2012}.
The main idea is to learn a word representation according to its context: the surrounding words \cite{Baroni2010}. 
The words are projected on a continuous space and those with similar context should be close in this multi-dimensional space.
When word vectors are available, a similarity between two words can be measured by a  metric such as a cosine similarity.

Using word-embeddings for machine translation evaluation is appealing since they can be used to compute similarity between words or phrases in the same language (monolingual embeddings capture intrinsically synonymy or morphological closeness) or in two different languages (bilingual embeddings allow to directly compute a distance between two sentences in different languages).
We use the \textit{MultiVec} \cite{MultiVecLREC2016} toolkit for computing and managing the continuous representations of texts.
It includes word2vec \cite{MikolovICLR2013}, paragraph vector \cite{paragraph-vector} and bilingual distributed representations \cite{luong_bilingual_2015} features.

\subsubsection{Use of vector representations in NLP evaluation}
\label{subsec:WEeval}

\newcite{Zou2013} proposed to use bilingual word embeddings to detect similarities for word alignment.
This information is used as an additional parameter in a phrase-based machine translation system.
\cite{Banchs2015} proposed to explore a metric funded on latent semantic analysis \cite{Salton1975} to extract semantic embeddings and measure the similarity between two sentences.
Finally, these word embeddings were used to enrich ROUGE, a metric for evaluating automatic summarization \cite{Ng2015}.

As far as MT evaluation is concerned, \newcite{metrics15rnn} proposed a metric based on neural network language models jointly with dependency trees to link an hypothesis to a reference.
Meanwhile, \newcite{metrics15docembed} proposed an approach to model document embeddings to predict  translation adequacy.

These works are close to ours but they propose metrics which need to be learned and optimized to a specific task or domain. 
In our work, we use word embeddings trained once and for all on a (large) general corpus.
Our detailed methodology to augment METEOR metric is presented in the next section.

\section{Augmented METEOR}
\label{sec:eval}
\subsection{Data and protocol}
We evaluate our augmented METEOR  through WMT14 framework (\textit{metrics} task \cite{Machacek2014}).
This framework enables us to estimate the correlation of proposed evaluation metric with human judgements for several machine translation outputs and several language pairs (English-French, English-German, English-Russian, and vice versa).
In our experiments, we use segment level Kendall's $\tau$ correlation coefficient, as proposed in WMT14 (based on systems ranking at sentence level by humans, compared to automatic metric ranking).

We augment METEOR in two ways: 
firstly, we replace the use of lexical resources by the use of word embeddings. In other words, we replace \textit{Stem} and \textit{Synonym} modules by our new \textit{Vector} module.
Secondly, we combine lexical resources and word embeddings by using jointly \textit{Stem}, \textit{Synonym}  and our \textit{Vector} module.

To summarize, the following variants of METEOR are evaluated: 
\begin{itemize}
 \item \textit{METEOR Baseline}: the METEOR score is estimated using \textit{Exact}, \textit{Stem}, \textit{Synonym} and \textit{Paraphrase} modules for English as a target language and \textit{Exact}, \textit{Stem} and \textit{Paraphrase} modules for other target languages,
 \item \textit{METEOR DBnary}: similar to \textit{METEOR Baseline} but \textit{Synonym} module is available for any target language since it uses DBnary resource instead of Wordnet, 
 \item \textit{METEOR Vector}: the \textit{Stem} and \textit{Synonym} modules are replaced by the \textit{Vector} module ;
 \item \textit{METEOR Baseline} $+$ \textit{Vector}: the \textit{METEOR Baseline} configuration is augmented with the \textit{Vector} module ;
 \item \textit{METEOR DBnary} $+$ \textit{Vector}: the \textit{METEOR DBnary} configuration is augmented with the \textit{Vector} module.
\end{itemize}

%
%
%
%

\subsection{\textit{METEOR DBnary}}

As mentioned in section \ref{subsec:meteor}, the \textit{Synonym} module of METEOR uses WordNet's synsets (117K entries for English).
As an alternative, we use another lexical resource: DBnary \cite{Serasset2012}, as proposed recently by \newcite{Elloumi2015}. This allows us to use \textit{Synonym} module for any target language: French, German, Spanish, Russian and English.

More precisely, synonym relations are extracted from DBnary using SPARQL request on the DBnary server\footnote{\url{http://kaiko.getalp.org/about-dbnary/online-access/}}.
We extract data for English, French, Russian and German languages. The extraction process outputs relations in the following format: $lemma$ $\rightarrow$ $Synonym$.
Then, these data are projected to the WordNet format used in METEOR code. This process gives an identifier (ID) for each lemma and builds a list of synonym IDs for each lemma such as: $ lemma~\rightarrow~ID\_Syn_1,~ID\_Syn_2,~ID\_Syn_3$.

The first two lines of Table \ref{tab:resDBnaryWE} compare  \textit{METEOR DBnary} and \textit{METEOR Baseline} for several French-English MT systems submitted to WMT14 \cite{Bojar2014}.

%
%

\textit{METEOR DBnary} improved the score by 0.7 points from \textit{METEOR Baseline}.
In other words, DBnary seems to match more synonyms than WordNet, despite the fact that WordNet is 3.3 time bigger than DBnary in English.
This could be due to the fact that WordNet has only 4 morpho-syntactic categories (Noun, Verbs, Adjectives and Adverbs) while DBnary has more morpho-syntactic categories.

\subsection{\textit{METEOR Vector}}

As mentioned in section \ref{subsec:WEeval}, we propose to replace lexical resources by word embeddings.
Word embeddings capture the context of the words. Consequently, similar word vectors may correspond to synonyms or morphological variants (see section \ref{sec:WE}).

%

\begin{table}[h]
\scriptsize
\centering
 \begin{tabular}{llrrr}
\multirow{2}{*}{Language} & \multirow{2}{*}{Corpora} & \multirow{2}{*}{\# of lines} & \# of source & \# of target \\
 & &  & words & words \\
 \hline
French--English & Europarl V7 $$+$$ news commentary V10  & 2.2 M & 67.2 M & 60.7 M  \\
German--English & Europarl V7 $$+$$ news commentary V10  & 2.1 M & 57.2 M & 59.7 M  \\
Russian--English & Common Crawl $$+$$ news commentary V10 $$+$$ Yandex & 2.0 M & 47.2 M & 50.3 M  \\
\hline
\end{tabular}
\caption{Bilingual corpora used to train the word embeddings for each language pair.}
\label{tab:corpora}
\end{table}

In our \textit{Vector} module, the matching between two words is done using a similarity score derived from the cosine similarity.
If the similarity score is higher than a threshold, the words are considered as matched (potential synonyms or morphological proximity). In our experiments, we evaluate using: (a) a default threshold fixed to 0.80 (b) an oracle threshold  obtained empirically on the WMT14 data set \cite{Machacek2014}.

Table \ref{tab:corpora} summarizes data used to train monolingual word embeddings and bilingual word embeddings.
These word embeddings were trained with a CBOW model, a vector size of 50 and a windows size $\pm$5 words, thanks to the MultiVec toolkit \cite{MultiVecLREC2016}.

\begin{table}[h]
\scriptsize
\centering
 \begin{tabular}{lcccccc}
 \hline

 \multirow{2}{*}{Metrics} & \multicolumn{5}{c}{Systems:}  \\
 & online A  & online B  & online C  & rbmt 1  & rbmt 4 \\
 \hline
 \textit{METEOR Baseline}  & 36.33 & 36.71 & 31.19 & 33.00 & 31.65 \\
 \textit{METEOR DBnary}  & 36.93 & 37.33 & 32.01 & 33.69 & 32.42 \\
 \textit{METEOR Vector} & 37.00 & 37.34 & 31.87 & 33.67 & 32.34 \\
 \textit{METEOR Baseline} $+$ \textit{Vector} & 37.08  &  37.40  &  31.96  &  33.75  &  32.45  \\
 \textit{METEOR DBnary} $+$ \textit{Vector} & 37.53  &  37.88  &  32.60  &  34.32  &  33.05  \\
 \hline

 \end{tabular}
  \caption{METEOR scores (all configurations) on the \textit{newstest} corpus of the  WMT14 translation evaluation task from French to English.}
 \label{tab:resDBnaryWE}
\end{table}

The results presented in table \ref{tab:resDBnaryWE} show that word embeddings (\textit{Vector} module) can efficiently replace lexical resources (\textit{Synonym} and \textit{Stem} modules) to match words in the translation hypothesis with those in the reference. In addition, their combination shows a good potential to match even more words between hypothesis and reference. 
In the next  section, we evaluate if the proposed versions of \textit{augmented} METEOR better correlate with human judgements.

\section{Correlations of Augmented METEOR with Human Judgements}
\label{sec:correlation}
\subsection{Results of different METEOR configurations}
In these experiments, we present results obtained with the \textit{Vector} module based on two threshold values: a default one (0.80) and an oracle one which maximizes the correlation with human judgement.

Table \ref{tab:correlationWMT14} presents the correlation scores obtained within the framework of WMT14 metrics task \cite{Machacek2014}\footnote{For better readability, we do not add standard deviations in the tables. These numbers will be, however, provided in supplementary material put on the paper web page (\url{https://github.com/cservan/METEOR-E/paper}).}.
The evaluation is done according to several translation tasks: from English to French (en--fr), German (en--de) and Russian (en--ru), and vice versa.
French, German and Russian as target languages represent a growing difficulty due to their morphology. English as target language allows to compare the lexical databases (Wordnet \textit{vs} DBnary).

\textbf{To English.} Firstly, when the translation direction is to English, we can observe that \textit{METEOR Baseline} and \textit{METEOR Vector} get equivalent results in average.
\textit{METEOR DBnary} also obtains similar results to \textit{METEOR Baseline}.
When we combine WordNet lexical resource and word embeddings (\textit{METEOR Baseline} $+$ \textit{Vector}), the reference score is increased by  $0,005$ points.
If the combination is done with DBnary's lexical data (\textit{METEOR DBnary} $+$ \textit{Vector}), the improvement is similar.

For \textit{Vector} module, optimization of the threshold slightly improves the average correlation. Combination of \textit{METEOR Baseline} $+$ \textit{Vector} or \textit{METEOR DBnary} $+$ \textit{Vector} improves by $0,002$ points when the threshold is optimized.

\textbf{From English.} Secondly, when the translation direction is from English, we can observe an improvement of the correlation score obtained with \textit{METEOR DBnary}, compared with \textit{METEOR Baseline}.
This is due to the fact that for French, German and Russian as target languages, \textit{METEOR Baseline} does not use any \textit{Synonym} module.
Our \textit{METEOR Vector} with the default threshold also gets better correlation scores compared to \textit{METEOR DBnary} ($+0.003$ points in average).
The combinations \textit{METEOR Baseline} $+$ \textit{Vector} and \textit{METEOR DBnary} $+$ \textit{Vector} further improve correlations with human judgements ($+0.001$ points in average).
Finally, when we use an oracle threshold for  \textit{Vector} module, improvements are  bigger and can reach  $0.013$ points in average, compared to \textit{METEOR Baseline}.

\begin{table}[h!]
\centering
\scriptsize
\begin{tabular}{lcccccc|cc}
\hline
Language pairs & \multicolumn{2}{c}{fr--en} & \multicolumn{2}{c}{de--en} & \multicolumn{2}{c}{ru--en} & \multicolumn{2}{c}{Average} \\
Metric & Threshold & $\tau$ & Threshold & $\tau$& Threshold & $\tau$ & Threshold & $\tau$ \\
\hline
\textit{METEOR Baseline} 			&  -- & 0.406 & -- & 0.334 & -- & 0.329 & -- & 0.356 \\
\textit{METEOR DBnary}  			&  -- & 0.408 & -- & 0.334 & -- & 0.328 & -- & 0.357 \\
\textit{METEOR Vector}   			&  0.80 & 0.407 & 0.80 & 0.332 & 0.80 & 0.328 & 0.80 & 0.356 \\
\textit{METEOR Baseline} $+$ \textit{Vector} 	&  0.80 & 0.407 & 0.80 & 0.343 & 0.80 & 0.332 & 0.80 & 0.361 \\
\textit{METEOR DBnary} $+$ \textit{Vector} 	&  0.80 & 0.407 & 0.80 & 0.337 & 0.80 & 0.338 & 0.80 & 0.361 \\
\hdashline
\textit{METEOR Vector}   			&  0.89 & 0.411 & 0.78 & 0.333 & 0.80 & 0.328 & 0.82 & 0.357 \\
\textit{METEOR Baseline} $+$ \textit{Vector} 	&  0.73 & 0.412 & 0.80 & 0.343 & 0.88 & 0.333 & 0.80 & 0.363 \\
\textit{METEOR DBnary} $+$ \textit{Vector} 	&  0.73 & 0.413 & 0.79 & 0.338 & 0.80 & 0.338 & 0.77 & 0.363 \\
\hline
\end{tabular}

\centering
\scriptsize
\begin{tabular}{lcccccc|ccc}
\hline
Language pairs & \multicolumn{2}{c}{en-fr} & \multicolumn{2}{c}{en-de} & \multicolumn{2}{c}{en-ru} & \multicolumn{2}{c}{Average} \\
Metric & Threshold & $\tau$ & Threshold & $\tau$& Threshold & $\tau$ & Threshold & $\tau$ \\
\hline
\textit{METEOR Baseline} 			& -- & 0.280 & -- & 0.238 & -- & 0.427 & -- & 0.315 \\
\textit{METEOR DBnary}  					& -- & 0.284 & -- & 0.239 & -- & 0.435 & -- & 0.319 \\
\textit{METEOR Vector}   			&  0.80 & 0.290 & 0.80 & 0.241 & 0.80 & 0.436 & 0.80 & 0.322\\
\textit{METEOR Baseline} $+$ \textit{Vector} &  0.80 & 0.288 & 0.80 & 0.241 & 0.80 & 0.440 & 0.80 & 0.323 \\
\textit{METEOR DBnary} $+$ \textit{Vector} 		&  0.80 & 0.289 & 0.80 & 0.242 & 0.80 & 0.439 & 0.80 & 0.323 \\
\hdashline
\textit{METEOR Vector}   			&  0.72 & 0.295 & 0.79 & 0.241 & 0.72 & 0.439 & 0.74 & 0.325 \\
\textit{METEOR Baseline} $+$ \textit{Vector} &  0.86 & 0.296 & 0.79 & 0.242 & 0.79 & 0.445 & 0.81 & 0.328 \\
\textit{METEOR DBnary} $+$ \textit{Vector} 		&  0.88 & 0.294 & 0.75 & 0.245 & 0.79 & 0.443 & 0.81 & 0.327\\
\hline

\end{tabular}

\caption{Correlation score at segment level between several METEOR configurations and human judgements (WMT14 framework). 
Scores obtained with the \textit{Vector} module are presented firstly with the default threshold (0.80) and secondly with the oracle threshold (under the dashed line).}
\label{tab:correlationWMT14}
\end{table}

\subsection{Investigating more embeddings configurations}

In the previous section, \textit{METEOR Vector} used a simple and monolingual word embedding configuration. This section investigates more configurations (monolingual and bilingual) to improve METEOR.

In this experiment, we focus only on \textit{METEOR Vector}. 
Indeed, the \textit{monolingual (baseline)} shown in table 6 corresponds to the line \textit{METEOR Vector} in Table  \ref{tab:correlationWMT14}.
Firstly, we propose to train our embeddings on bitexts (Table \ref{tab:corpora}) using \textit{bivec} approach \cite{luong_bilingual_2015}.
We also try to train pseudo-bilingual embeddings on a pseudo bitext with target language text and POS tags (see an example in Table \ref{ex:biPOS}).
\begin{table}[h!]
\centering
\scriptsize

\begin{tabular}{ccc}
\hline
\\
madam president , on a point of order .
& $\Leftrightarrow$ & 
NOUN NOUN PUNCT ADP DET NOUN ADP NOUN PUNCT \\
\\
\hline
\end{tabular}
\caption{Example of bitext where the target side is replaced by POS.}
\label{ex:biPOS}
\end{table}
The main idea is to strongly link words with their syntactic class when learning word embeddings. 
We call this kind of model \textit{pseudo-bilingual with POS}.
In the same way, we train bilingual models called \textit{pseudo-bilingual with lemmas}, where the POS tags are replaced by lemmas. 
In addition, we also learn word embeddings with lemmas only and bilingual models with lemmas only.

\begin{table}[h!]
\centering
\scriptsize
\begin{tabular}{lcccccc}
\hline
\multirow{2}{*}{Models:}  & \textit{monolingual}   & \multirow{2}{*}{\textit{bilingual}}  & \textit{pseudo-bilingual}  & \textit{pseudo-bilingual}  & \textit{monolingual}  & \textit{bilingual} \\
& \textit{(baseline)}   & & \textit{with POS}  & \textit{with lemmas}  & \textit{(lemmas)}  & \textit{(lemmas)} \\
\hline
To English  & 	 0.356  &  0.354  &  0.355  &  0.354  &  0.357  &  0.357\\
From English & 0.322  & 0.322  & 0.320  & 0.325  & 0.324  & 0.318\\

\hline
\end{tabular}
\caption{Average correlation score at segment level for \textit{METEOR Vector} with several training configurations of word embeddings with the default threshold (0.80).} 
\label{tab:EMBComp}
\end{table}

In the Table \ref{tab:EMBComp}, we compare several training configuration of the word embeddings through the same protocol as previous section (only average correlations are reported while the detailed results will be provided as supplementary material on the paper web page).
When we observe the average results, the \textit{bilingual} embeddings seem not to be as efficient as the monolingual baseline. 
The pseudo-bilingual approaches with POS and Lemmas obtained slightly the same results as the monolingual baseline regarding all the configurations we have.
Finally, the monolingual model learned on lemmas (instead of words) tends to be slightly better when the translation direction is to English. However, this trend should be confirmed in a future investigation.

\subsection{Discussion}

The correlation scores obtained with the enriched metric tend to suggest that  distributed representations are as powerful as lexico-semantic resources for  automatic MT evaluation. Furthermore, vector representations can bring  additional information, and they are definitely useful when no lexical resource is available in the target language.

Considering the average correlation scores obtained, the configurations \textit{METEOR Vector} and  \textit{METEOR DBnary} are comparable, except on German language, for which \textit{METEOR Vector} obtained a better correlation score.
On the other hand, when we combine lexical data with  \textit{Vector} module (\textit{METEOR DBnary} $+$ \textit {Vector}), we observe a small increase of the correlation score, in particular when threshold is tuned, which suggests a tunable version of METEOR.

Finally, several embeddings variants were trained but it seems that  monolingual models are efficient enough for the specific task (MT evaluation) considered here.
 
\subsubsection{Examples}

To illustrate the word matching obtained by our versions of METEOR, we analyze two examples from the evaluation data set.
In these examples, we present the alignments obtained with \textit{METEOR DBnary} $+$ \textit{Vector}.

\begin{table}[h!]
\centering
\scriptsize
\begin{tabular}{ll}
\hline
\\
 hypothesis: & \dbox{je}	\dbox{pense}	qu'	il	est	concevable	que	ces	données	soient	\fbox{employées}	\fbox{pour}	le	bénéfice	mutuel	.  \\
 Reference: & \dbox{j'}	\dbox{estime}	qu'	il	est	concevable	que	ces	données	soient	\fbox{utilisées}	\fbox{dans}	leur	intérêt	mutuel	. \\
 \\
 \hline
\end{tabular}
\caption{First example from the system \textit{rbmt 1} evaluated with the combination \textit{METEOR DBnary} $+$ \textit{Vector}. The relations detected with the lexical resource DBnary are framed in continuous line while those obtained thanks to the distributed representations are framed in dotted line.}
\label{tab:ex1}
\end{table}
The example presented in table \ref{tab:ex1} shows \textit{rbmt 1} system output submitted during the WMT14 translation task.
\textit{METEOR baseline} found only alignments for words with the same surface forms (\textit{``qu' '', ``il'', ``est''}, etc. -- these forms are found identical thanks to the \textit{Exact} module and are not highlighted here). 
The \textit{Synonym} module based on DBnary makes it possible to find a correspondence between words \textit{``employées''} -- \textit{``utilisées''} and \textit{``pour''} -- \textit{``dans''}. 
Lastly, \textit{Vector} module indicates that words \textit{``pense''} and \textit{``estime''} are contextually closed, just as the words \textit{``je''} and \textit{``j'''}. 
When the example is only evaluated with \textit{METEOR Vector}, words \textit{``employées''} and \textit{``utilisées''} are also paired with the default threshold  (0.80). 
On the other hand, the words  \textit{``bénéfice''} and \textit{``intérêt''} are paired by the module \textit{Vector} only if the  decision threshold is lowered to 0.75.

\begin{table}[h]
\centering
\scriptsize
\begin{tabular}{ll}
\hline
\\
 hypothesis: & le	\fbox{créateur}	de	SAS	disait	il	\fbox{faisait}	un	genre	\dbox{du}	feuilleton	géopolitique	.\\
 Reference: & le	\fbox{père}	de	SAS	disait	\fbox{faire}	un	genre	\dbox{de}	feuilleton	géopolitique	.\\
 \\
 \hline
\end{tabular}
\caption{Another example scored with the combination \textit{METEOR DBnary} $+$ \textit{Vector}. }
\label{tab:ex2}
\end{table}

In the second example presented in table \ref{tab:ex2}, the hypothesis is provided by \textit{rbmt 4} system.
As in the previous example, the correspondences found with \textit{Synonym} module based on DBnary (framed by one continuous line) are supplemented by those found by \textit{Vector} module (dotted line):
\textit{Synonym} module found \textit{``créateur''} -- \textit{``père''} and \textit{``faisait''} -- \textit{``faire''};
while \textit{``du''} and \textit{``de''} are aligned thanks to \textit{Vector} module. 

These examples illustrate the complementarity between  lexical resources and word embeddings for sentence similarity detection. 
Word vectors can enable to match important words (like \textit{``pense''} and \textit{``estime''} in our first example), but also empty words (like \textit{``du''} et \textit{``de''} in our second example).

\subsubsection{Limitations of Word Embeddings}

So far, we did not deal with Out-Of-Vocabulary (OOV) words in \textit{METEOR Vector}. 
By OOV we mean words that do not have a vector representation because they were not found in the training corpus for word embeddings.
In that case, no matching can occur between the word in the hypothesis and words in reference. 
Consequently, it might be interesting to carefully select the training corpus for word vectors so that it will be close enough to the machine translation outputs to evaluate. This could be considered in future works.
 
\section{Conclusion and Perspectives}
\label{sec:concl}

In this paper, we proposed to compare text similarity measures based on vector representations  with similarity measures based on lexico-semantic resources.
Our work was applied to machine translation evaluation and we extended an existing evaluation metric called METEOR.
Our experiments have shown that word vector representations can be useful when no lexical resource  is available in the target language.
Moreover, it seems that these representations can bring complementary information in addition to lexical resources (experiments done for French, English, German and Russian as target languages). 

Our future works on this topic will focus on the use of phrase embeddings to complement the \textit{Paraphrase} module of METEOR. We also plan to introduce a \textit{syntax flavor} in our \textit{Vector} module by weighting the cosine distances differently according to the morpho-syntactic category of the words. 
Finally, we will study the adaptation of our approach to other metrics such as TER-Plus, for instance.

The tool, the data and the models presented in this paper will be put online\footnote{\url{https://github.com/cservan/METEOR-E}} to facilitate reproducibility of the experiments we carried out.

\section*{Acknowledgements}

%

This work was supported by the KEHATH project funded by the French National Agency for Research (ANR) under the grant number ANR-14-CE24-0016-03.

\bibliographystyle{acl}
\bibliography{biblioCOLING2016}

\end{document}